\documentclass{article}

\usepackage{arxiv}

\usepackage[utf8]{inputenc} 
\usepackage[T1]{fontenc}    
\usepackage{hyperref}       
\usepackage{url}            
\usepackage{booktabs}       
\usepackage{amsfonts}       
\usepackage{nicefrac}       
\usepackage{microtype}      
\usepackage{lipsum}
\usepackage{xr}
\usepackage{placeins}  

\usepackage{longtable}
\usepackage[utf8]{inputenc} 
\usepackage[T1]{fontenc}    
\usepackage{url}            
\usepackage{booktabs}       
\usepackage{amsfonts}       
\usepackage{nicefrac}       
\usepackage{microtype}      
\usepackage{lipsum}
\usepackage{fancyhdr}       
\usepackage{graphicx}       
\graphicspath{{media/}}     
\usepackage{tabularx}
\usepackage{listings} 
\usepackage{fancyvrb} 

\usepackage[numbers]{natbib}
\usepackage{amsmath,amssymb}
\usepackage{amsthm}
\usepackage{mathtools}
\usepackage{xspace}
\usepackage[noend]{algorithmic}
\usepackage[ruled,vlined]{algorithm2e}
\usepackage{enumitem}

\usepackage{pdfpages}

\usepackage{chngcntr}
\usepackage{mdframed} 
\usepackage{caption}

\usepackage{color}
\usepackage{xcolor}

\usepackage{subfigure}
\usepackage[most]{tcolorbox}
\usepackage[frozencache=true, finalizecache=false, cachedir=./minted-cache]{minted}

\usepackage{float}
\usepackage{alltt}
\usepackage{soul}
\usepackage{fancyvrb}
\usepackage{multirow}

\usepackage{csquotes}

\newcounter{textbox}
\usepackage{mdframed} 

\usepackage{listings} 

\tcbset{
  custombox/.style={
    width=474.18663pt,
    top=3pt,
    fonttitle=\bfseries\small\sffamily, 
    colback=white,
    colframe=black,
    colbacktitle=black,
    boxrule=0.5pt, 
    colback=white, 
    enhanced,
    rounded corners, 
    center,
    boxed title style={
            sharp corners,
            size=small,
            colframe=blue!75!black, 
        },
    attach boxed title to top left={yshift=-0.1in,xshift=0.15in},
    boxed title style={boxrule=0pt,colframe=white,},
  }
}

\newtcolorbox{LLMbox}[2][]{custombox,title=#2,#1}

\tcbset{
  customboxmultipage/.style={
    width=474.18663pt,
    top=3pt,
    fonttitle=\bfseries\small\sffamily, 
    colback=white,
    breakable,
    colframe=black,
    colbacktitle=black,
    boxrule=0.5pt, 
    colback=white, 
    enhanced,
    rounded corners, 
    center,
    boxed title style={
            sharp corners,
            size=small,
            colframe=blue!75!black, 
        },
    attach boxed title to top left={yshift=-0.1in,xshift=0.15in},
    boxed title style={boxrule=0pt,colframe=white,},
  }
}
\newtcolorbox{LLMboxmultipage}[2][]{customboxmultipage,title=#2,#1}

\newtcbox{\mybox}[1][green]{on line,
arc=0pt,outer arc=0pt,colback=#1!10!white,colframe=#1!50!black,
boxsep=0pt,left=0pt,right=0pt,top=0pt,bottom=0pt,
boxrule=0pt,bottomrule=0pt,toprule=0pt}

\definecolor{aigold}{RGB}{244,210, 1} 
\definecolor{aigreen}{RGB}{210,244,211} 

\sethlcolor{aigreen}

\definecolor{aired}{RGB}{255,180,181}

\definecolor{lightred}{rgb}{1,0.9,0.9} 

\title{\bf{AutomataGPT: Forecasting and Ruleset Inference for Two-Dimensional Cellular Automata}
}

\author{
  Jaime A. Berkovich\textsuperscript{1,2} \quad
  Noah S. David\textsuperscript{1} \quad
  Markus J. Buehler\textsuperscript{1,3,4,5,\#} \\
  \\
  \textsuperscript{1}Laboratory for Atomistic and Molecular Mechanics (LAMM)\\
  \textsuperscript{2}Department of Materials Science and Engineering\\
  \textsuperscript{3}Department of Civil and Environmental Engineering, \\
  \textsuperscript{4}Department of Mechanical Engineering, \\
  \textsuperscript{5}Center for Computational Science and Engineering, \\
  Schwarzman College of Computing, \\
  Massachusetts Institute of Technology, Cambridge, MA 02139, USA \\ \\
    \textsuperscript{\#}Corresponding author: 
  \texttt{mbuehler@MIT.EDU}
}

\begin{document}
\maketitle
\begin{abstract}

Cellular automata (CA) provide a minimal formalism for investigating how simple local interactions generate rich spatiotemporal behavior in domains as diverse as traffic flow, ecology, tissue morphogenesis and crystal growth. However, automatically discovering the local update rules for a given phenomenon and using them for quantitative prediction remains challenging. Here we present AutomataGPT, a decoder-only transformer pretrained on ~1 million simulated trajectories that span 100 distinct two-dimensional binary deterministic CA rules on toroidal grids. When evaluated on previously unseen rules drawn from the same CA family, AutomataGPT attains 98.5\% perfect one-step forecasts and reconstructs the governing update rule with up to 96 \% functional (application) accuracy and 82 \% exact rule-matrix match. These results demonstrate that large-scale pretraining over wider regions of rule space yields substantial generalization in both the forward (state forecasting) and inverse (rule inference) problems, without hand-crafted priors.
By showing that transformer models can faithfully infer and execute CA dynamics from data alone, our work lays the groundwork for abstracting real-world dynamical phenomena into data-efficient CA surrogates—opening avenues in biology, tissue engineering, physics and AI-driven scientific discovery.
\end{abstract}


\section{Main}\label{sec:main}
\begin{figure}
    \centering
    \includegraphics[width=1\linewidth]{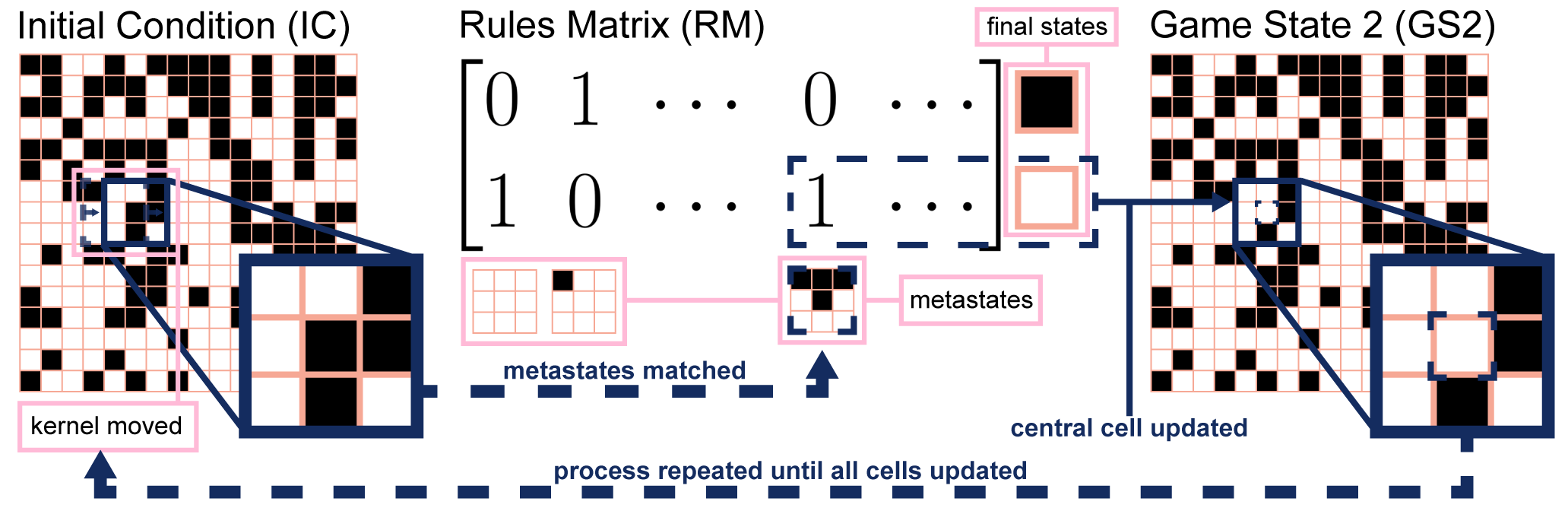}
    \caption{Illustration of the process of computing one future global state (or `game state') for a 2D binary deterministic \(r=1\) cellular automaton, based on an inputted initial \(16\times16\) binary grid representing an initial global state/initial condition (IC) and 2D binary array representing local state transition rules, referred to here as a `rules matrix' (RM). Local states are updated based on a \(3\times3\) kernel (moving window) that scans the IC and matches observed cell-neighborhood pairs (metastates) with their corresponding columns in the RM. For a given metastate's RM column, a row is selected if it contains a 1, indicating 100\% probability of state transition. The selected row corresponds to the state to which the cell in the center of the kernel is reassigned. Note: for this study, we consider toroidal grids (2D grids with periodic boundary conditions).}
    \label{fig:rules-app}
\end{figure}

Cellular automata (CA) are a class of algorithms that consist of an array of cells, each having a single `state' out of a finite set of possible states, with recursive, local update rules governing state transitions at discrete time intervals (Fig.~\ref{fig:rules-app}). These algorithms have been of particular interest for computer scientists and natural scientists alike, due to their ability to exhibit complex, emergent phenomena from specific, simple rules~\cite{kari_theory_2005}. In the context of computer science, there are CA algorithms -- both one-dimensional (1D) and two-dimensional (2D) -- that have been found to be Turing complete (computationally universal)~\cite{kari_theory_2005}. Researchers have historically proposed variations of 2D CA as novel parallel computational architectures, as each cell in a 2D CA grid may be updated independently, and therefore in parallel~\cite{mitchell_computation_1998}. In the natural sciences, CA offer computationally efficient frameworks for simulating emergent/coarse-grained dynamics in physical systems across a broad spectrum of lengthscales, timescales, and substrates~\cite{kari_theory_2005,zhao_cellular_2020,hernandez_encinas_simulation_2007,kier_modeling_2005,ding2011modeling}.

An intriguing feature of cellular automata (CA) in interdisciplinary contexts is the very quality that makes them challenging to formally characterize. Namely, this is the fact that CA often exhibit highly complex or chaotic, `computationally irreducible' behavior, implying that any prediction of the state \(S\) of a CA system \(n\) timesteps in the future must take at least \(n\) discrete computational steps to produce -- there are no lossless `shortcuts' or analytical representations of the system as a function of \(n\) and \(S\)~\cite{Wolfram2002,zwirn_unpredictability_2013,ufr_de_physique_lied_universite_paris_7_cmla_ens_cachan_france_and_ihpst_cnrs_france_computational_2015}. Thus, while CA systems have served well as pedagogical tools for qualitatively illustrating the emergence of complexity from simple underlying rules, they remain widely unutilized as quantitative predictive models for physical systems. With regard to this, Springer and Kenyon (2020)~\cite{springer_its_2020} showed that it often takes much larger convolutional neural networks (CNNs) than theoretically necessary to reliably learn the state-transition rules of the 2D CA system `Conway's Game of Life' (Life)~\cite{gardner_mathematical_1970}. Results such as these have led to the common assumption that it is unlikely that AI will be able to perfectly predict the \(n^{\text{th}}\) state of a particular CA system based on an initial condition (IC).

However, even if a given CA system is likely irreducible, there is nothing stopping us from slowly evolving state transition rules to generate some target emergent behavior. Neural cellular automata~\cite{mordvintsev_growing_2020} (NCA) have also emerged as a subfield focused on synthesizing the tunability of neural networks with the collective dynamics of cellular automata. NCA have been shown to enable dynamical CA systems to  generate emergent visual patterns with tunable steady states~\cite{richardson_learning_2024}. The key attribute of NCA is that each cell in a 2D grid is encoded as a vector and the global update rule is itself a neural network (CNN~\cite{mordvintsev_growing_2020}, Attention Layer~\cite{tesfaldet_attention-based_2022}, etc.) which intakes cells' local neighborhood cell vectors and outputs incremental updates to cells' states. Using this framework, the update rule can be tuned iteratively through backpropagation. While this field shows real promise for the future of CA-based dynamic mimicry of synthetic and real-world systems, the fact that the state transition rules in NCA are themselves neural networks fundamentally limits their \textit{interpretability}. In essence, we cannot easily `peer into' these systems to find what leads to their unique, dynamical evolution -- their state transition rules are, in effect, `black boxes.' 

Still, one does not need to use neural networks as update rules to develop a framework for iterative tuning of emergent behavior. In the case of spacetime-inhomogeneous cellular automata~\cite{wolfram_whats_2024}, randomly `mutating' the placement of CA rules within `spacetime' and keeping only those mutations which advance a particular performance metric while rejecting the rest has been shown to eventually lead to a configuration engendering a desired behavior; though, this strategy is undoubtedly under-researched and 
may turn out to be limited depending on the particular performance metric, rule space, and grid size.

On the other side of the computer science discipline, transformer models have gained widespread popularity across both computer science and the natural sciences for their domain-agnostic learning capabilities and favorable scaling behaviors~\cite{vaswani_attention_2017,kaplan_scaling_2020}, including but not limited to genomic analysis~\cite{liu_metagene-1_2025}, meteorology and climate prediction~\cite{nguyen_scaling_2023, alerskans_transformer_2022, saleem_stc-vit_2024}, protein structure prediction~\cite{moussad_transformative_2023}, strategic decision-making in games~\cite{xu_language_2024, hu_survey_2024, liu_large_2024}, computational linguistics and text generation~\cite{gillioz_overview_2020}, audio and speech recognition~\cite{latif_transformers_2023, mamatov_speech_2021, li_neural_2019}, computer vision and image analysis~\cite{berroukham_vision_2023}, and materials science~\cite{buehler_fieldperceiver_2022, buehler_mechgpt_2024,buehler_preflexor_2024_updated,GraphPreflexorBuehler_2025}, among others.

We have recently shown~\cite{berkovich_lifegpt_2024_updated} that generative pretrained transformer models (GPTs) are capable of learning the state-transition rules of CA from raw data. In the resulting model, LifeGPT, we showed that a decoder-only transformer model was able to learn to apply the update rules of Conway's Game of Life on a \(32\times32\) toroidal grid, with no prior knowledge of grid topology, presenting near-perfect accuracy. Following this, Burtsev (2024)~\cite{burtsev_learning_2024} used elementary (1D) CA to investigate the ability for GPTs to predict long-term behavior in complex systems, notably finding that increasing model depth, including intermediate future states, and explicitly including CA ruleset inference into training loss improved performance. These results support the hypothesis we presented in our previous work~\cite{berkovich_lifegpt_2024_updated} that CA rules could function as an internal `world model,' as posited by Ha and Schmidhuber (2018)~\cite{ha_world_2018}: a simplified construct of the system at hand that serves to guide the model in the `correct direction.'

Similarly meshing transformer models with artificial life, Kumar \textit{et al.} (2024) proposed Automated Search for Artificial Life (ASAL)~\cite{kumar_automating_2024}, a paradigm incorporating vision-language foundation models (FMs) for discovering and taxonomizing artificial life (ALife) simulations across multiple substrates (Boids~\cite{reynolds_flocks_1987}, Particle Life~\cite{tom_mohr_how_2022,ventrella_clusters_nodate}, Life-like CA~\cite{pena_life_2021,wojtowicz_cellular_nodate}, Lenia~\cite{hong_kong_lenia_2019,faldor_toward_2024}, and NCA~\cite{mordvintsev_growing_2020}). In this work, vision language FMs were used within a larger genetic algorithm to generate representations of ALife simulations (spacetimes) in order to match natural language prompts, maximize simulation `open-endedness,' and maximize simulation novelty relative to explore/taxonomize simulation-space.

Both Kumar \textit{et al.} (2024)~\cite{kumar_automating_2024} and Burtsev (2024)~\cite{burtsev_learning_2024} showed that training transformer models on large sets of rules leads to adequate model generalization for emergent dynamics in ALife systems. However, Burtsev focused on 1D CA forecasting while Kumar \textit{et al.} focused on exploration and taxonomy of `simulation-space' for a variety of 2D substrates, including Life-like CA. Neither work investigated forecasting the evolution of 2D CA, nor did they investigate whether rules could be inferred/extracted by an AI system, given some input ALife time-evolution/CA orbit.

\begin{figure}
    \centering
    \includegraphics[width=\linewidth]{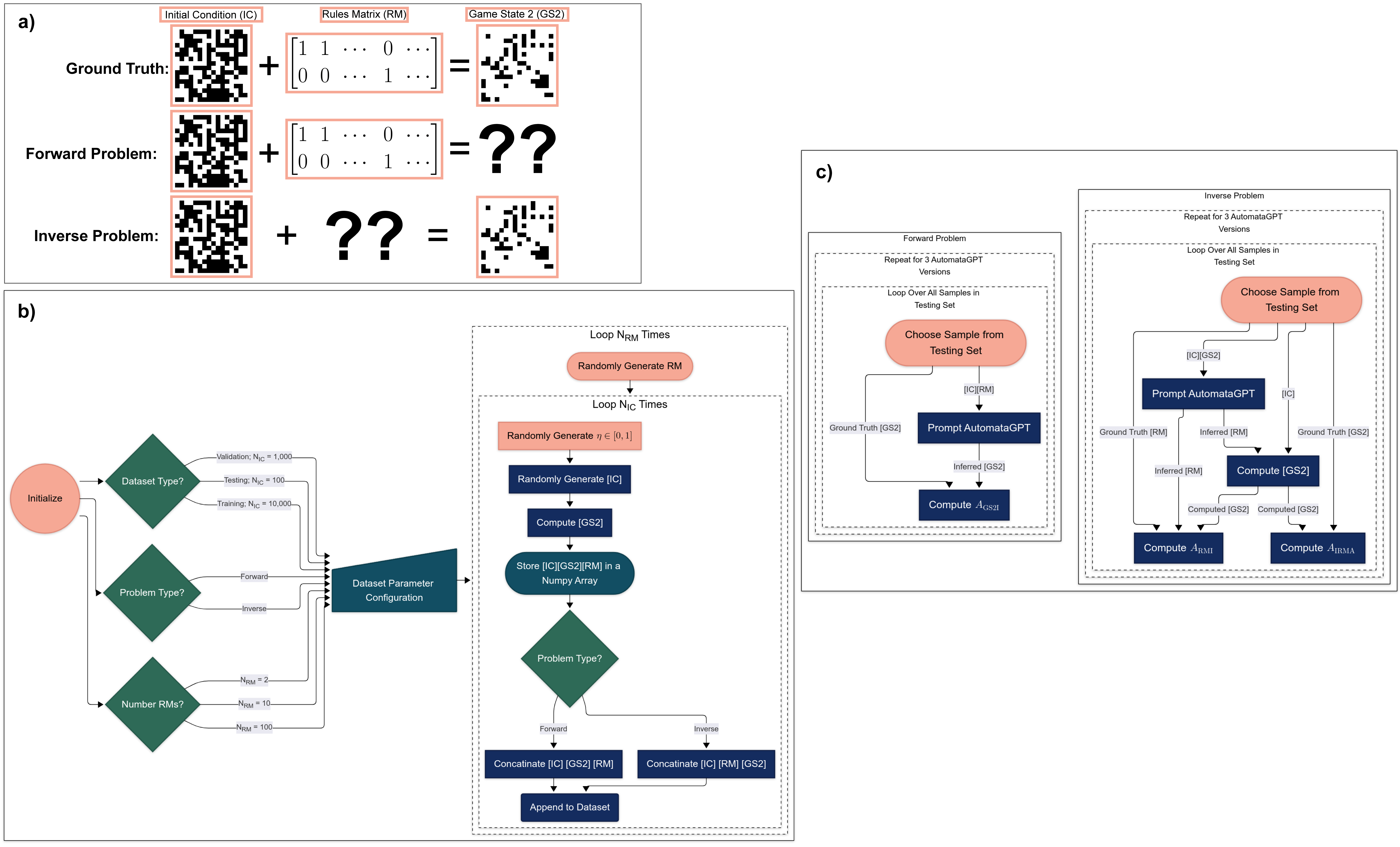}
    \caption{\textbf{a)} Illustration of the forward and inverse problems investigated in this study, with respect to the ground truth. The ground truth shows the standard relationship between the IC, RM, and GS2 for a given CA system sample: the rules defined by the RM are applied to the IC to compute GS2. Question marks represent the unknowns (GS2 or RM) that AutomataGPT is tasked with inferring for each problem. \textbf{b)} Dataset generation procedure for datasets used to train, validate, and test AutomataGPT. For each \(N_{\text{RM}}\) value, a unique version of AutomataGPT was trained from scratch  -- three models per problem for a total of six unique versions of AutomataGPT (see Section~\ref{sec:model_versions}). It should be noted that in addition to the training, validation, and testing sets generated for each AutomataGPT version, one additional testing set (\(N_\text{RM}=100; N_\text{IC}=2\)) was generated for characterizing performance \textit{across all} AutomataGPT versions. \textbf{c)} Experimental procedure for characterizing the performance of versions of AutomataGPT for both the forward and inverse problems (using the same \(N_\text{RM}=100; N_\text{IC}=2\) testing set).}
    \label{fig:dataset-performance}
\end{figure}

This prior work leaves open the following questions: (1) Does the improved \textit{forecasting} (Fig.~\ref{fig:dataset-performance}a) performance of GPTs enabled by explicitly encoding rules in training data (as shown by Burtsev) extend to a setting for systems with increased dimensionalities (2D) and topological complexities (toroidal grid)? (2) How well can GPTs solve an inverse problem (Fig.~\ref{fig:dataset-performance}a) in which successive states of a CA system are given and the model is tasked with \textit{inferring rules} that fit the system's dynamics?

We believe answering these questions is critical to the development of novel, CA-based, physical simulation techniques. Considering that morphological development in multicellular organisms in governed by the global emergent dynamics of local cellular interactions~\cite{fratzl_natures_2007,xavier_da_silveira_dos_santos_single_2019}, generating CA rulesets representative of these biological systems using GPTs prompted with empirical data could be an effective technique for forecasting time-evolution, which would be highly beneficial for fields such as biology and tissue engineering~\cite{yang_learning_2024,bagherpour_application_2024,kundrat_comprehensive_2010,ko_computational_2022, fletcher_seven_2022,sharpe_computer_2017}. Furthermore, many non-biological natural phenomena are emergent effects stemming from the interactions of many particles or otherwise discrete units of matter~\cite{kivelson_defining_2016,ponge_emergent_2005,schmickl_how_2016,kesic_systems_2016}. To this end, generating CA rules from empirical data might help us bridge the explanatory gap between length scales across many systems and scientific domains. 

Thus, we propose \textbf{AutomataGPT}, a decoder-only GPT model that we trained in two distinct manners to (1) apply update rules to 2D binary deterministic CA with a \(r=1\) Moore neighborhood~\cite{weisstein_moore_nodate} and (2) infer CA rules given an initial condition (IC) and a subsequent game state (GS2). For each problem, we train three versions of AutomataGPT from scratch on different sets of CA rules (different numbers of RMs: \(N_\text{RM}\in\{2,10,100\}\)) and on both the forward and inverse problems (Fig.~\ref{fig:dataset-performance}a). Our experiments investigate how the size of the rule space (the number of RMs -- see Section~\ref{sec:rmform} --  used to construct a CA dataset -- see Fig.~\ref{fig:dataset-performance}b) used for training impacts accuracy for both problems (Fig.~\ref{fig:dataset-performance}c), with a focus on RM inference performance as a function of test sample RM dissimilarity relative to RMs in training data. In doing so, we benchmark generalization capabilities and inductive bias across different AutomatGPT versions.

While our study focuses on demonstrating the feasibility of ruleset inference and forecasting within 2D CA systems, the broader implications of this work extend to real-world applications in biology, physics, and other natural sciences. These applications, though beyond the scope of this paper, highlight the potential for CA-based approaches to bridge explanatory gaps across scales and domains.

\paragraph{Paper Outline}
We begin by investigating versions of AutomataGPT trained on the forward problem. We define an accuracy metric, interpret data from experiments testing three versions trained on different training sets, and offer questions for future research. We then investigate versions of AutomataGPT trained on the inverse problem with distinct training data, defining two new accuracy metrics and providing data from our experiments. We then interpret our findings, commenting on the relationship between training data and AutomataGPT's generalization capabilities, inductive bias, and creativity. We conclude by discussing implications for modeling real-world dynamical systems and potential next-steps.

\FloatBarrier  
\section{Results and Discussion}
\subsection{Forward Problem}
We defined the forward problem as: ``If given a flattened representation of the RM (see sections \ref{sec:rmform} and \ref{sec:rmg}) and an IC, can AutomataGPT accurately generate the following state of the CA grid in accordance with the RM and IC, `game state 2' (GS2)?'' (See Fig.~\ref{fig:rules-app}.)

\begin{figure}[hbt!]
    \centering
    \includegraphics[width=\linewidth]{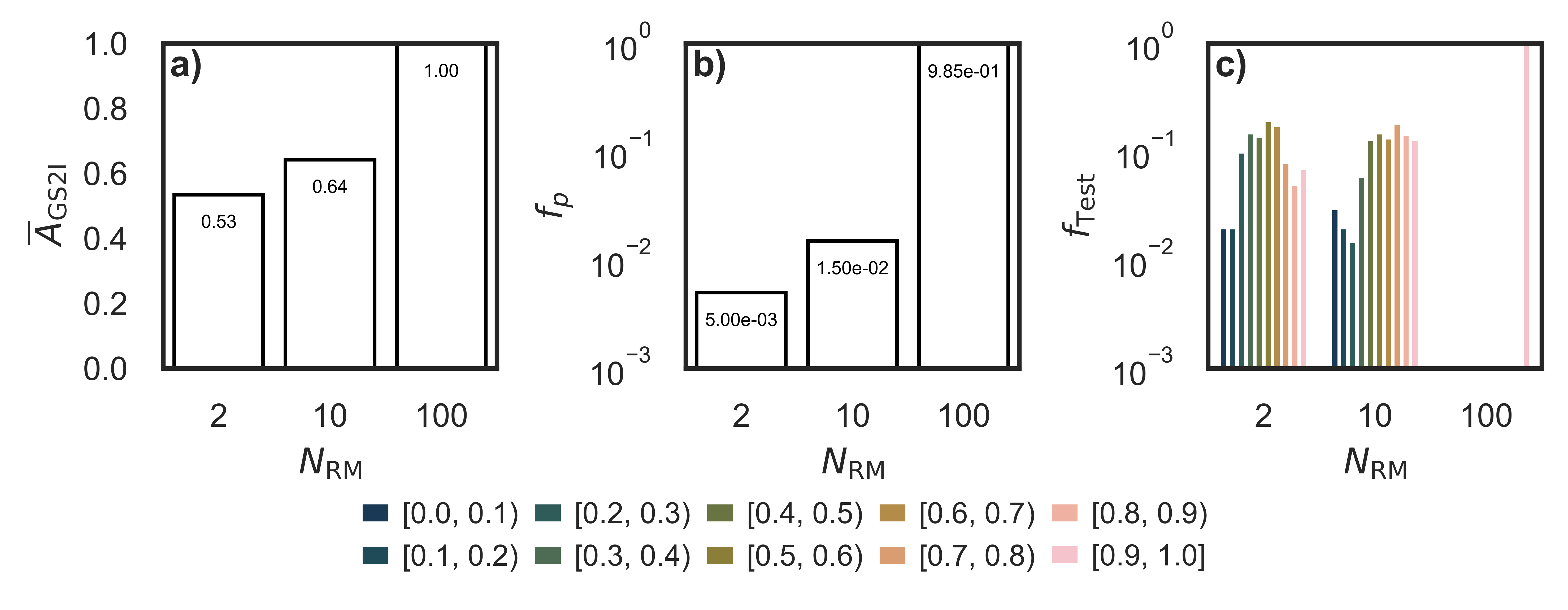}
    \caption{Accuracy increases significantly with more extensive training on rulesets, demonstrating AutomataGPT’s generalization from 2 to 100 rules. \textbf{a)} Bar chart illustrating the change in AutomataGPT's accuracy on the forward problem averaged over the entire testing set (\(\overline{A}_{\text{GS2I}}\)) versus the number of RMs used for training (\(N_\text{RM}\)). \textbf{b)} Bar chart depicting the affect of \(N_\text{RM}\) on the fraction of perfect GS2 inferences across the testing set (\(f_p\)). \textbf{c)} \(A_{\text{GS2I}}\) histogram for models trained with varying \(N_\text{RM}\), where \(f_\text{Test}\) is the fraction of samples in the testing set. The grayscale legend at the bottom of the figure denotes \(A_{\text{GS2I}}\) ranges for the histograms. Charts \textbf{a-c)} all indicate that, in the case of the forward problem, as AutomataGPT was trained on larger rule spaces, its inference accuracy consistently increased, reaching near-perfect accuracy for \(N_\text{RM}=100\). This suggests that for \(N_\text{RM}\), AutomataGPT was effectively trained to be a CA computer, capable of applying any RM (corresponding to a 2D binary deterministic automaton) to any \(16\times16\) binary IC to generate a near-perfect GS2 inference. }
    \label{fig:fwd_acc_nrm}
\end{figure}

Fig.~\ref{fig:fwd_acc_nrm} illustrates the change in AutomataGPT's inference accuracy for the forward problem as the model is trained from scratch on an increasing number of RMs. For each version of AutomataGPT, we quantify forecasting accuracy using the accuracy of GS2 inference (\(A_{\text{GS2I}}\)), where \(A_{\text{GS2I}}\) is calculated independently for each sample in the testing set, as is defined as follows:

\begin{equation}\label{eq:GS2I}
A_{\text{GS2I}} = \frac{\gamma_c}{\gamma_t}
\end{equation}
where \(\gamma_c\) is the number of correctly \textit{inferred} GS2 tokens and \(\gamma_t=16^2=256\) is the total number of tokens comprising each GS2. Furthermore, \(\overline{A}_{\text{GS2I}}\) is the average accuracy of GS2 inference (averaged across all samples in the testing set):
\begin{equation}
    \overline{A}_{\text{GS2I}} = \frac{1}{N_S}\sum_{i=0}^{N_S-1}A_{\text{GS2I},i},
\end{equation}
where \(A_{\text{GS2I},i}\) is the accuracy of GS2 inference for the \(i^{\text{th}}\) sample in the testing set and \(N_S\) is the number of samples in the testing set (\(N_S=200\)).

We found that \(\overline{A}_{\text{GS2I}}\) increased as AutomataGPT was trained on increasing \(N_{\text{RM}}\) (Fig.~\ref{fig:fwd_acc_nrm}a). This suggests that training GPTs on larger sections of rule space improves their ability to become generally programmable for running CA. In other words, when a GPT model is trained many examples of CA governed only by a \textit{few} distinct RMs, it likely finds non-intuitive token relationships that do not generalize for unseen RMs; however, when a GPT model is trained on many examples of CA with \textit{many} corresponding RMs, it successfully learns a `near-correct' method of applying the inputted RM onto the inputted IC to yield a GS2, becoming \textit{rules-agnostic}. Still, in our experiment for \(N_{\text{RM}}=100\), the model was not perfect, yielding perfect GS2 inferences \(98.5\%\) of the time, inferring 3 out of 200 GS2 imperfectly (Fig.~\ref{fig:fwd_acc_nrm}b). Nevertheless, the difference is stark -- the \(N_{\text{RM}}=100\) AutomataGPT version is near-perfect while being trained on less than \(0.04\%\) of all possible rules (see section~\ref{sec:rules_calc} and equation~\ref{eq:100RMsfrac}). \(N_{\text{RM}}=100\) was trained using 1e4 ICs per RM, coming out to one million `games' in the training set. We can safely assume that the vast majority of these games have unique ICs, due to the massive IC configuration space: \(2^{16\times16} \approx 1.15 \text{e}77\) binary ICs for a \(16\times16\) grid. This means that the \(N_\text{RM}=100\) AutomataGPT version was trained on only \(100\%\times\frac{1\text{e}6}{1.15\text{e}77}\approx 8.64\times 10^{-70}\%\) of all possible ICs. This demonstrates the incredible ability of \textit{small} (relative to the state-of-the-art LLMs) GPTs to generalize without ultra-large training sets.

Given these results, we suspect that the observed increase in inference accuracy would continue to approach perfection if \(N_{\text{RM}}\) were further increased. Furthermore, it is shown in Fig.~\ref{fig:fwd_acc_nrm}c that for \(N_{\text{RM}}=200\), \(100\%\) of samples tested yielded \(A_{\text{GS2I}}\geq0.9\) (\(\geq90\%\) forecasting accuracy). These results exist in contrast to the relatively flat accuracy distributions for \(N_{\text{RM}}=2\) and \(N_{\text{RM}}=10\) -- in fact, \(N_{\text{RM}}=10\) showed marginal improvement over \(N_{\text{RM}}=2\), suggesting that increasing \(N_{\text{RM}}\) leads to \textit{gradual} model generalization improvements, as \textit{opposed} to there existing a critical \(N_{\text{RM}}\) for which the model undergoes a `phase transition.'

These results suggest that it is generally possible for a decoder-only transformer model to be trained such that it becomes a near-perfect, generally-programmable, 2D binary deterministic CA computer, and that increasing the breadth of the rules in its training data will improve its single-timestep forecasting accuracy, even if the number of rules it is trained only comprises a small fraction of the whole rule space.  Still, it is difficult to say with much certainty at what \(N_{\text{RM}}\), if any, AutomataGPT would reach \(100\%\) forecasting accuracy -- it may be that GPTs will often get stuck in local minima instead of effectively generalizing for this type of problem. Further research is required to address these questions.

\subsection{Inverse Problem} 
We defined the inverse problem as: ``Given an IC and GS2 for a CA system, can AutomataGPT accurately infer the RM?'' (See Fig.~\ref{fig:rules-app}.)

\begin{figure}[hbt!]
    \centering
    \includegraphics[width=\linewidth]{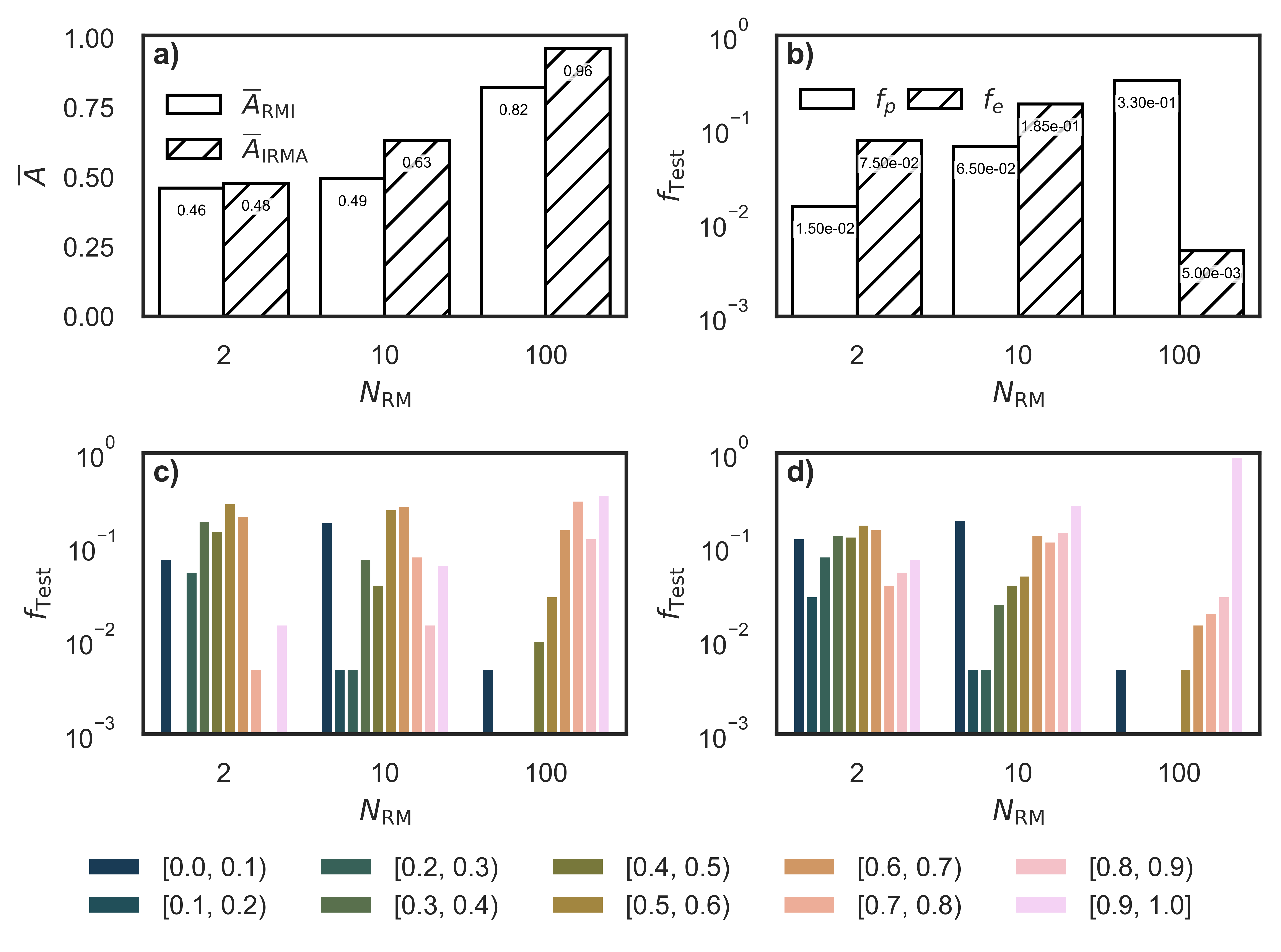}
    \caption{\textbf{a)} Bar chart illustrating the change in inverse problem inference accuracy averaged over the testing set (\(\overline{A}\)) for two distinct accuracy metrics (Rules Matrix Inference Accuracy -- \(\overline{A}_{\text{RMI}}\) -- and Inferred Rules Matrix Application Accuracy -- \(\overline{A}_{\text{IRMA}}\)) versus the number of RMs used for training (\(N_\text{RM}\)). This chart shows that as AutomataGPT is trained on a larger rule space, its inferences become more accurate overall. That fact that \(\overline{A}_{\text{RMI}}<\overline{A}_{\text{IRMA}}\forall N_\text{RM}\) suggests that even relatively inaccurate RM inferences can yield somewhat more accurate GS2 computations when applied a given sample's IC. Moreover, the widening gap between \(\overline{A}_{\text{RMI}}\) and \(\overline{A}_{\text{IRMA}}\) as \(N_\text{RM}\) increases suggests that training on larger rule spaces enables AutomataGPT to infer more accurate \textit{degenerate solutions} (RMs that are not the same as ground truth but still yield somewhat accurate GS2s when applied to ICs). \textbf{b)} Bar chart depicting the affect of \(N_\text{RM}\) on the fraction of perfect RM inferences across the testing set (\(f_p\)) and the fraction of errors (inferred RMs that are illogical) across the testing set (\(f_e\)). It is observed that training on a larger rule space enables fewer illogical RM inferences and more perfect RM inferences. \textbf{c)} \(\overline{A}_{\text{RMI}}\) histogram for models trained with varying \(N_\text{RM}\). \textbf{d)} \(\overline{A}_{\text{IRMA}}\) histogram for models trained with varying \(N_\text{RM}\). Charts \textbf{c)} and \textbf{d)} indicate the same trends as \textbf{a)} but in more detail, showing the fraction of test samples belonging to each \(\overline{A}\) range. Observing these data, it is clear that the improvement gains brought on by training on a larger rule space are more substantial when considering degenerate solutions. One can interpret these findings as evidence of the model becoming more `creative' as it is trained on larger rule spaces.} 
    \label{fig:acc_nrm}
\end{figure}

\begin{figure}
    \centering
    \includegraphics[width=\linewidth]{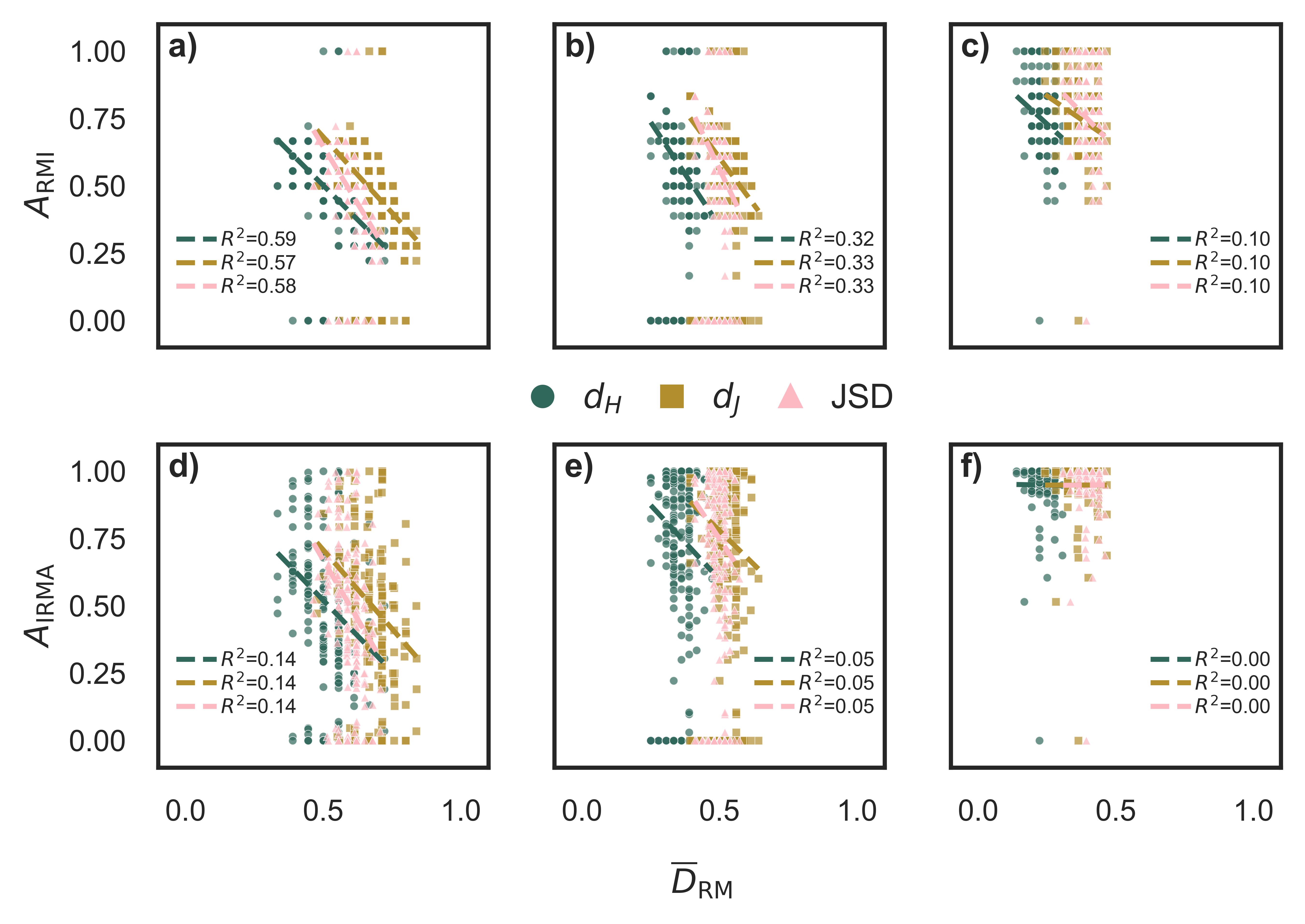}
    \caption{Plots depicting the capabilities of varying versions of AutomataGPT to infer RMs (the inverse problem) for a testing set with samples generated using RMs not present in any training data. Specifically, model accuracy vs. RM difference averaged across the closest two RMs in the training set (\(\overline{D}_{\text{RM}}\)) is plotted for varying difference metrics, models, and accuracy metrics. The \(\overline{D}_{\text{RM}}\) metrics used were Hamming distance (\(d_H\)), Jaccard distance (\(d_J\)), and Jensen-Shannon Divergence (JSD). \textbf{a)} Rules Matrix Inference Accuracy (\(A_{\text{RMI}}\)) vs. \(\overline{D}_{\text{RM}}\) for \(N_{\text{RM}}=2\). \textbf{b)} \(A_{\text{RMI}}\) vs. \(\overline{D}_{\text{RM}}\) for \(N_{\text{RM}}=10\). \textbf{c)} \(A_{\text{RMI}}\) vs. \(\overline{D}_{\text{RM}}\) for \(N_{\text{RM}}=100\). \textbf{d)} Inferred Rules Matrix Application Accuracy (\(A_{\text{IRMA}}\)) vs. (\(\overline{D}_{\text{RM}}\)) for \(N_{\text{RM}}=2\). \textbf{e)} \(A_{\text{IRMA}}\) vs. \(\overline{D}_{\text{RM}}\) for \(N_{\text{RM}}=10\). \textbf{f)} \(A_{\text{IRMA}}\) vs. \(\overline{D}_{\text{RM}}\) for \(N_{\text{RM}}=100\). The data shown across plots \textbf{a-f)} provide insight into the inductive bias of AutomataGPT versions depending on the number of RMs used for training. In particular, a decreasing \(R^2\) value across all \(\overline{D}_{\text{RM}}\)s as \(N_{\text{RM}}\) increases indicates that as AutomataGPT learns from larger rule spaces, it more effectively generalizes across rules it has never seen before, with the rules that is \textit{has} seen playing an increasingly insignificant role in shaping its responses. Furthermore, the fact that, for \(N_{\text{RM}}=100\), \(R^2=0.00\) for \(A_{\text{IRMA}}\) while \(R^2=0.10\) for \(A_{\text{RMI}}\) suggests that this AutomataGPT version is more likely to generate degenerate solutions for the inverse problem when a given test sample has been generated with an RM very different from RMs in the model's training set (creativity is prompted by `unusual systems') -- notably, such degenerate solutions are just as likely to reflect the dynamics of the test sample system as non-degenerate solutions. Note: Accuracies of 0 and 1 were excluded from the regressions, ensuring that regressions only correlated the accuracies of \textit{imperfect-but-still-logical} RM inferences with \(\overline{D}_{\text{RM}}\).}
    \label{fig:inv_inf}
\end{figure}

Fig.s~\ref{fig:acc_nrm} and~\ref{fig:inv_inf} illustrate the change in AutomataGPT's inference accuracy for the inverse problem as it was repeatedly trained from scratch on an increasing number of RMs. For each version of AutomataGPT, we quantify inference accuracy using \textit{two} metrics.

The first accuracy metric is \(A_{\text{RMI}}\) (Rules Matrix Inference Accuracy), defined as:
\begin{equation}\label{eq:RMI}
A_{\text{RMI}} = \begin{cases}
1, & \text{if the inferred RM yields the exact ground truth GS2} \\
0, & \text{if the inferred RM is illogical}\\
\frac{\rho_c}{\rho_t}, & \text{otherwise}
\end{cases}
\end{equation}
where \(\rho_c\) is the number of correctly inferred tokens in the RM, and \(\rho_t=36\) is the total number of tokens comprising each RM. \(A_{\text{RMI}}\) was defined this way in order to not penalize our model for finding alternative RMs equally capable of \textit{perfectly} describing the CA system in the testing set.

The second accuracy metric is \(A_{\text{IRMA}}\) (Inferred Rules Matrix Application Accuracy), defined as:
\begin{equation}
     A_{\text{IRMA}} = \begin{cases}
     0, & \text{if the inferred RM is illogical}
     \\
     \frac{\gamma_c}{\gamma_t}, & \text{otherwise}
     \end{cases}
\end{equation}
where \(\gamma_c\) is the number of correctly computed tokens for GS2 when \textit{applying} the inferred RM to the IC, and \(\gamma_t=16^2=256\) is the total number of tokens comprising each GS2. \(A_{\text{IRMA}}\) captures the ability of our model to infer imperfect RMs that still reasonably capture the behavior of the CA system in the testing set.

We also considered the average values of these two accuracy metrics, across all samples in the testing set:
\begin{equation}\label{eq:AvRMI}
     \overline{A}_{\text{RMI}} = \frac{1}{N_S}\sum_{i=0}^{N_S-1}A_{\text{RMI},i},
\end{equation}
\begin{equation}\label{eq:AvIRMA}
     \overline{A}_{\text{IRMA}} = \frac{1}{N_S}\sum_{i=0}^{N_S-1}A_{\text{IRMA},i},
\end{equation}
where \(A_{\text{RMI},i}\) and \(A_{\text{IRMA},i}\) were the accuracies for the \(i^{\text{th}}\) samples in the testing set and \(N_S\) was the number of samples in the testing set (\(N_S=200\)).

We found that for both accuracy metrics, the average accuracy (\(\overline{A}\)) increased as AutomataGPT was trained with higher \(N_{\text{RM}}\) (Fig. \ref{fig:acc_nrm}a). This is evidence that increasing the number of RMs in the training set results in an increased ability for our model to generalize across rule space. The fraction of samples for which AutomataGPT perfectly inferred RMs (including inferring degenerate RMs) also steadily increased with \(N_{\text{RM}}\)(Fig. \ref{fig:acc_nrm}b).

Also, for \(\overline{A}_{\text{RMI}}\), the fraction of samples in the testing set leading to the inference of an illogical RM (where probabilities of state transition for a given metastate do not add to 1), which we termed the `error fraction' (\(f_e\)), initially increased from \(N_{\text{RM}}=2\) to (See Fig.~\ref{fig:rules-app}.), but decreased by over an order of magnitude from \(N_{\text{RM}}=10\) to \(N_{\text{RM}}=100\) (Fig. \ref{fig:acc_nrm}b). This suggests that for a sufficiently great number of RMs AutomataGPT learns to infer mostly \textit{logical} RMs. We suspect that for even larger \(N_{\text{RM}}\)s, this effect would be intensified, and AutomataGPT's illogical inferences would become exceedingly rare. Another indication that increasing the number of RMs was beneficial overall to model performance was that AutomataGPT's accuracy distribution improved favorably with higher \(N_{\text{RM}}\) when considering both \(A_{\text{RMI}}\) and \(A_{\text{IRMA}}\) (Fig. \ref{fig:acc_nrm}c).

While these results initially lent hope to the notion that a large enough training set might allow AutomataGPT to accurately infer rules for \textit{any} 2D binary deterministic CA system, there was still the possibility that the diversity of the RMs in AutomataGPT's training data had inevitably limited its inference capabilities. It seemed hypothetically feasible that samples in the testing set corresponding to CA systems that were computed with RMs  similar to those used to create the training set would always result in more accurate RM inferences. \textit{If} this was found to be true, then it would be likely that AutomataGPT has significant inductive bias.

To test for this possibility, we defined \(\overline{D}_{\text{RM}}\), representing an information-theoretic distance quantifying the average dissimilarity between the testing sample's RM and the `closest' \textit{two} RMs in AutomataGPT's training set. While we refer to \(\overline{D}_{\text{RM}}\) generally, we specifically utilized three established dissimilarity metrics in order to remove metric-induced bias from our results: (1) \textit{Hamming distance} -- \(d_H\)~\cite{hamming_error_1950}; (2) \textit{Jaccard distance} (or one minus the \textit{Jaccard index}, which is also known as the \textit{Tanimoto coefficient}) -- \(d_J\)~\cite{jaccard_distribution_1912,tanimoto_elementary_1958}; (3) \textit{Jensen-Shannon divergence} (also known as the \textit{information radius}) -- JSD~\cite{nielsen_jensenshannon_2019}.

As AutomataGPT was trained on an increasing number of rulesets, inverse problem accuracies (both \(A_{\text{RMI}}\) and \(A_{\text{IRMA}}\)) became less correlated with \(\overline{D}_{\text{RM}}\). This was shown by the linear regressions in Fig.~\ref{fig:inv_inf}, which approximated the function \(\overline{A}_{\text{RMI}}(\overline{D}_{\text{RM}})\) \(\forall \overline{A}_{\text{RMI}} \notin \{0,1\}\) in Fig.s~\ref{fig:inv_inf}a-c and \(\overline{A}_{\text{IRMA}}(\overline{D}_{\text{RM}})\) \(\forall \overline{A}_{\text{IRMA}} \notin \{0,1\}\) in Fig.s~\ref{fig:inv_inf}d-f. Accuracies of 0 and 1 were excluded from the regressions since the former reflected illogically inferred RMs (not interpretable as adjacency matrices) and the latter reflected perfectly inferred RMs (matching ground truth).

This trend held true across all \(\overline{D}_{\text{RM}}\)s (\(d_H\), \(d_J\), and JSD). This suggested that as versions of  AutomataGPT were trained on increasing numbers of rulesets, they became increasingly general -- the rules governing the test sample system became less correlated with inference accuracy.

Furthermore, the \(R^2\) values across all \(\overline{D}_{\text{RM}}\)s decreased for all models when switching the chosen accuracy metric from \(A_{\text{RMI}}\) to \(A_{\text{IRMA}}\) (Fig. \ref{fig:inv_inf}). In particular, for \(N_{\text{RM}}=100\), \(R^2\) dropped from 0.10 to 0.00 (see Figs.~\ref{fig:inv_inf}c and~\ref{fig:inv_inf}f), which showed that while AutomataGPT was more likely to infer RMs close to ground truth for smaller \(\overline{D}_{\text{RM}}\), the extent to which these inferred RMs were accurate reflections of the underlying dynamics of the test sample system was not correlated with \(\overline{D}_{\text{RM}}\). Fundamentally, this means that AutomataGPT is more likely to come up with `creative' (divergent inference) solutions to the inverse problem if the test sample RM was far from all RMs in its training set. Importantly, creativity did \textit{not} trade-off with accuracy. For \(N_{\text{RM}}=100\), AutomataGPT was just as likely to accurately infer RMs that reliably reproduced CA dynamics for systems with low \(\overline{D}_{\text{RM}}\) as for those with high \(\overline{D}_{\text{RM}}\) (Fig. \ref{fig:inv_inf}f).

Potentially, the reason that AutomataGPT finds degenerate solutions (RMs that do \textit{not} match the ground truth RM but still work well to compute GS2 when applied to the IC) to inverse problems is because it lacks enough information to select only ground truth RMs. If this where found to be true, then if given more time-evolution data (longer orbits) for CA systems, we might expect AutomataGPT to become better at inferring ground-truth RMs, as the space of degenerate solutions would shrink. Another related question is: ``How does Automata GPT's \textit{sampling temperature} affect creativity and accuracy with regard to the inverse problem?'' We deem these inquiries to be beyond the scope of this paper and worthy of further research. 

The implications of our work are not limited to the realm of CA research. AutomataGPT could be the first step towards a new framework for modeling real-world complex systems and systems exhibiting emergence. We show that the state-transition rules for 2D-binary-deterministic CA can be inferred using AutomataGPT, and that inference accuracy is highly influenced by the breadth of the rules in the training set. Future models trained on CA systems having more states, more spatial dimensions, and larger grids might be able to learn rules across many physical systems, so long as such systems may be effectively coarse-grained using discrete cells. Such versions of AutomataGPT for forecasting physical systems would have a \textit{three-fold} advantage over existing AI-enabled techniques. 
\begin{itemize}
    \item [1.] \textbf{Use of Synthetic-only Training Data.} Since such models would only require examples of time-evolution for CA systems, datasets could be purely synthetic, eliminating the need for expensive real-world data collection.
    \item [2.] \textbf{Computational Efficiency.} Due to the high parallelizability of CA algorithms, physical simulations represented entirely with CA rules (inferred with future AutomataGPTs) could be efficiently scaled-up for larger systems.
    \item [3.] \textbf{Simulation Interpretability.} Finding relationships between CA and PDEs could enable a two-way simulation paradigm in which inferred CA rules could assist researchers with identifying new PDEs, conserved quantities, or other mathematical formalisms to advance understandings for a plethora of physical systems. In fact, Omohundro (1984)~\cite{omohundro_modelling_1984,yang_cellular_2010} describes a method for converting any 2D nine-neighbor square-lattice CA system into a corresponding system of partial differential equations (PDEs). Future work could feasibly synthesize similar methods with AutomataGPT-based methods to quickly define PDEs for nearly any 2D dynamical system, unlocking a world of dynamical systems analysis across a broad spectrum of scientific disciplines. Alternatively, CA rules could be interpreted intuitively by a human researcher, depending on the particular system at hand, potentially enabling the discovery of new analytic frameworks.
\end{itemize}

The success of this future research direction will depend on how effectively CA are generally able to simulate complex systems, but we are optimistic considering the wide range of behavior exhibited by CA of differing dimensionalities, state-possibilities, and rules. Another potential bottleneck for such a paradigm could be training costs. While GPTs have been shown to scale well with large training sets and context lengths~\cite{openai_gpt-4_2024,vaswani_attention_2017,kaplan_scaling_2020}, and have architectures which are highly conducive to training parallelization using GPUs~\cite{vaswani_attention_2017}, it is not clear exactly how large future versions of AutomataGPT would have to be to capture the complexity present across a plethora of physical systems. Answering such a question will require benchmarking various AutomataGPTs on a range of existing physical systems.

\FloatBarrier
\section{Conclusion}
We demonstrate a novel framework, AutomataGPT, for forecasting time-evolution (forward problem) and inferring rules (inverse problem) for 2D-binary-deterministic CA systems with toroidal boundary conditions. We observe that increasing the number of distinct rules used to create the training set significantly improves model performance on both the forward and inverse problems. Our findings suggest significant future advancements in artificial intelligence and physical modeling. AutomataGPT, trained exclusively on \textit{synthetic-only} data, accurately forecasts and infers rules from previously unseen 2D CA systems. 
Additionally, AutomataGPT shows stark improvements in performance and reductions in inductive bias for systems outside of its training set after being trained on increasingly wide sections of rule space. 
We hypothesize this framework could infer rules describing real-world systems without requiring large empirical datasets. Such a paradigm might dramatically reduce the costs (both monetary and temporal) of training AI models for physical prediction tasks. We believe it is crucial that future work be done to investigate the scaling behavior of AutomataGPT-like models, so as to determine the feasibility of training such models using larger CA rule spaces with more degrees of freedom (larger grid sizes, more cell states, and higher dimensionalities). If there exists enough latent complexity within large rule spaces, then future models might be able to learn to extract meaningful rules from a wide range of systems across varying length scales, so long as empirical time-evolution data can be provided using a prompt. The implications of AutomataGPT extend beyond CA, hinting at broader, transdisciplinary applications for which GPT-based architectures could be used not only to infer the behavior of systems, but to extract meaning in a way that advances human understanding, paving the path for human-AI symbiosis for scientific advancement. In that vein, AutomataGPT exemplifies a promising direction in bridging symbolic rule-based systems and subsymbolic neural architectures. The key concept put forth here is that by enabling interpretable rule inference alongside flexible forecasting, it paves the way for hybrid models that combine the transparency of cellular automata with the expressive power of large language models. Our findings support the idea that AI models need not merely fit or predict empirical data; they can also recover underlying generative mechanisms, akin to discovering governing equations. This suggests a path toward AI-native scientific frameworks in which models do not just simulate, but understand and re-express the rules of complex systems.

\section{Materials and Methods}
Codes and data are available at \href{https://github.com/lamm-mit/AutomataGPT}{https://github.com/lamm-mit/AutomataGPT}.

\subsection{Model Architecture and Training Hardware}
AutomataGPT was constructed in python using \href{https://github.com/lucidrains/x-transformers}{x-transformers} \cite{wang_lucidrainsx-transformers_2024}. The models in this study were trained on a workstation equipped with a high-end CUDA-compatible GPU (RTX A4000, NVidia, Santa Clara, CA, USA).

\subsection{Training}

\subsubsection{Hyperparameters}\label{sec:hyper}
Hyperparameters were initially selected heuristically for optimal performance, as the GPU primarily used for training (RTX A4000, NVidia, Santa Clara, CA, USA) had 16 GB of VRAM. Unless otherwise stated, all instances of AutomataGPT used the following set of hyperparameters during training, as described in Table~\ref{tab:hyper}.
\begin{table}[!htbp]
  \caption{AutomataGPT's hyperparameters.}
  \centering
  \begin{tabular}{ll}
    \toprule
    \textbf{Hyperparameter}      & \textbf{Value}         \\
    \midrule
    num\_tokens             & 22                    \\
    max\_seq\_len           & 555                   \\
    dim (\(d_{\mathrm{model}}\))                     & 256                    \\
    depth (\(N_{\mathrm{layers}}\))                   & 6                     \\
    heads (\(h\))                   & 4                      \\
    RPE        & \texttt{True}          \\
    Flash attention             & \texttt{True}          \\
    Optimizer               & Adam
    \\
    Learning rate          & 1e-4
    \\
    mask\_prob              & 0.15
    \\
    Loss function           & Cross entropy loss (CLE)
    \\
    Batch size              & 50
    \\
    Gradient accumulation period    & 1
    \\
    Training data IC ordering & broad-entropy
    \\
    Warm-up Scheduler & linear
    \\
    \bottomrule
    \label{tab:hyper}
  \end{tabular}
\end{table}
\subsubsection{Model Versions}\label{sec:model_versions}
\begin{table}[htb!]
\centering
\caption{Summary of all AutomataGPT versions featured in this study, detailing their key characteristics and problem type.}
\label{tab:model_summary}
\resizebox{\columnwidth}{!}{
\begin{tabular}{lccccccc}
\toprule
\textbf{Problem Type} & \(N_{\text{RM}}\) & \textbf{Model Name} & \textbf{Grid Size} & \textbf{Grid Topology} & \textbf{Deterministic CA?} & \textbf{Neighborhood} & \textbf{\# Cell States} \\
\midrule
Forward & 2   & AutomataGPT-2DBD-2rf  & $16 \times 16$ & 2D Toroidal & Yes & Moore (\(r=1\)) & 2 \\
Forward & 10  & AutomataGPT-2DBD-10rf & $16 \times 16$ & 2D Toroidal & Yes & Moore (\(r=1\)) & 2 \\
Forward & 100 & AutomataGPT-2DBD-100rf & $16 \times 16$ & 2D Toroidal & Yes & Moore (\(r=1\)) & 2 \\
Inverse & 2   & AutomataGPT-2DBD-2ri  & $16 \times 16$ & 2D Toroidal & Yes & Moore (\(r=1\)) & 2 \\
Inverse & 10  & AutomataGPT-2DBD-10ri & $16 \times 16$ & 2D Toroidal & Yes & Moore (\(r=1\)) & 2 \\
Inverse & 100 & AutomataGPT-2DBD-100ri & $16 \times 16$ & 2D Toroidal & Yes & Moore (\(r=1\)) & 2 \\
\bottomrule
\end{tabular}
}
\end{table}

\subsubsection{Training Loop}\label{sec:training_loop}
AutomataGPT-2DBD-2rf, -10rf, -2ri, and -10ri models were trained for 50 epochs, and AutomataGPT-2DBD-100rf and -100ri were trained for 20 epochs. For each batch within each epoch, training loss was calculated, followed by subsequent backpropagation. Then validation loss was calculated without calculating gradients, since this loss was not used for weight adjustment. After each epoch of training, AutomataGPT was tasked with inference -- which either required predicting the GS2 or the RM, depending on the model version/problem type (see section~\ref{sec:model_versions}) -- and accuracies were calculated across all samples in the testing set. For each batch, the corresponding epoch, global step, training loss, validation loss, testing accuracy, and elapsed time were logged.

\subsection{Datasets}
\subsubsection{Data Generation Overview}\label{sec:data-gen-overview}
To generate training sets, validation sets, and testing sets, a custom python script was used. First binary, deterministic RMs were stochastically generated. Then, IC game-states were generated stochastically as a 2D, 16 $\times$ 16 \verb+numpy+~\cite{harris_array_2020} arrays. Then the corresponding GS2 for every previously generated IC, based on each corresponding RM, was calculated. Lastly, each IC-GS2-RM triad was stored in a \texttt{numpy} array. When training or testing models, these numpy arrays were reformatted into lists of strings, with the relative ordering of the ICs, RMs, and GS2s depending on if the dataset was intended for the forward problem or inverse problem (see Section ~\ref{sec:instruction_tuning} and Fig.~\ref{fig:dataset-performance}). For the purpose of model performance characterization across all versions of AutomatGPT, a single testing set was used with dataset parameters \(N_\text{RM}=100; N_\text{IC}=2\).

\subsubsection{Data Topology}
Each untokenized sample in the training data (or validation/testing data) was represented as a 1D string where the rules-matrix, IC-matrix, and GS2-matrix were flattened and concatenated in varying orders, depending on the training task.

\subsubsection{RM Formalization}\label{sec:rmform}

\begin{figure}[!ht]
    \centering
    \includegraphics[width=\linewidth]{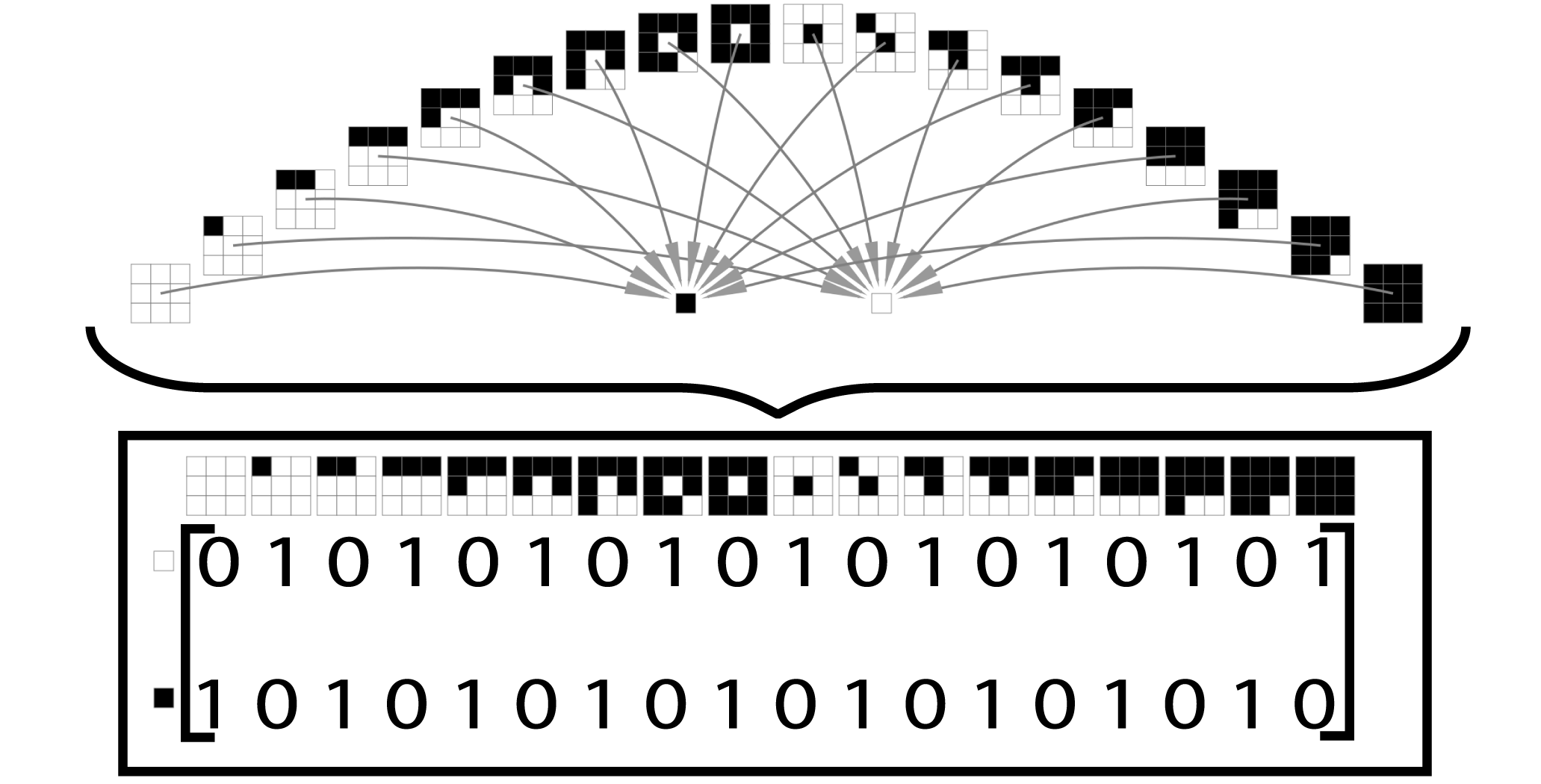}
    \caption{Transformation of a graph representation of state transition in 2D binary CA to a 2D binary adjacency matrix. We refer to the latter mathematical object as a `rules matrix.'}
    \label{fig:graph_to_matrix}
 \end{figure}

As shown in Fig. \ref{fig:graph_to_matrix}, an unweighted directional graph representing state transition for binary, deterministic CA may be losslessly represented as a binary adjacency matrix~\cite{weisstein_adjacency_nodate}. The rows of this matrix correspond to the final state of the cell undergoing transition, and the columns represent the possible initial `metastates' of the cell. We define a metastate as a \(3\times 3\) binary matrix representing the state of a cell along with its local (\(r=1\)) Moore neighborhood~\cite{weisstein_moore_nodate}. In other words, the metastate is a combination of the state of a cell and all of its neighbors. We assume metastates are permutation invariant, meaning the orientation of neighboring states does not change the identity of the metastate so long as the net number of neighboring cells in each possible state remains unchanged. Since binary (\(r=1\)) 2D automata have 18 metastates and 2 individual cell states we defined RMs as \(2\times 18\) 2D arrays.

\paragraph{Calculating the Total Number of Possible RMs.}\label{sec:rules_calc}

In a 2D deterministic binary (isotropic/count-based) CA with a Moore neighborhood, each cell can be in one of two possible states: \(0\) or \(1\).  The next state of the cell depends on its current state and the number of neighbors in state~\(1\). Because there are \(8\) neighbors, the neighbor count can range from \(0\) to \(8\), giving us:

\begin{equation}
\text{Possible neighbor counts} 
\;=\; 
9.
\end{equation}

Also considering the cell’s current state (two possibilities: \(0\) or \(1\)) brings the total to:

\begin{equation}
\text{Total metastates} 
\;=\; 
2 
\;\times\; 
9
\;=\;
18.
\end{equation}

A \emph{deterministic} rules matrix (RM) specifies the next state (either \(0\) or \(1\)) for \emph{each} of these 18 configurations.  Hence, for every metastate there are 2 possible next states, and thus the total number of possible deterministic RMs is \(2^{18}\). Therefore, for a 2D binary CA (\(r=1\) Moore neighborhood and non-directional rules), there are exactly \(2^{18}\) distinct RMs.

100 RMs, for instance, represents only a tiny fraction of the entire rule space:
\begin{equation}
    \text{Percentage of total rule space for 100 RMs} = 100\% \times \frac{100}{2^{18}}\cong 0.03814\%\approx 0.04\%.\label{eq:100RMsfrac}
\end{equation}

\subsubsection{RM Generation}\label{sec:rmg}
For the present work, we generate rule matrices (RMs) in a \emph{deterministic} manner by choosing exactly one next state for each \emph{metastate}. While our Python script also offers a stochastic mode, it was not used in this study. In addition, each RM is labeled with an integer \emph{Rules Matrix ID} (RMID), yet in the current implementation the ID is \emph{not} used to select the next states. Instead, these are chosen randomly at runtime.

\paragraph{Deterministic Rule Assignment.}
For each column (metastate) \(j\in\{1,\dots,c\}\) we \emph{randomly} choose a next state (in \(\{0,\dots,b-1\}\)) via \texttt{random.randint(0,\,b-1)}. Formally, for each column \(j\),
\[
\text{RM}[k,j]
\;=\;
\begin{cases}
1 & \text{if }k = \texttt{rand\_state}_{j}, \\
0 & \text{otherwise},
\end{cases}
\]
where \(\texttt{rand\_state}_{j}\) is a random integer from \([0,\dots,b-1]\). As a result, each metastate has \emph{exactly one} valid next state, but the choice of which one is random. Although this is still a fully deterministic rule (each metastate maps to a single next state), it is drawn stochastically at the time of generation.

\paragraph{Stochastic Mode (Not Used).}
Our code also supports a purely \emph{stochastic} mode, in which each column can encode a probability distribution over the possible next states. However, all results in this paper exclusively use the deterministic mode described above. (We anticipate that the stochastic mode will be useful for future studies on stochastic CA.)

\subsubsection{IC Generation}\label{sec:icg}
The initial states of each cellular automaton are generated through a stochastic process involving two main steps: random order parameter generation and cell-by-cell state assignment.

Firstly, for each cellular automaton, we generate a set of random order parameters that define a probability distribution over the possible cell states. Specifically, we generate $b-1$ random numbers (where $b$ is the base of the automaton), sort them in ascending order, and extend this list by including $0$ at the beginning and $1$ at the end. This results in a sequence:

\[
p_0 = 0 < p_1 < p_2 < \cdots < p_{n-1} < p_n = 1.
\]

The order parameters are defined as the absolute differences between consecutive elements:

\[
\Delta p_k = p_{k+1} - p_k, \quad \text{for } k = 0 \text{ to } n-1.
\]

$\Delta p_k$ represents the probabilities of assigning state $k$ to a cell, such that
$\sum_{k=0}^{n-1} \Delta p_k = 1$.

Secondly, each cell in the cellular automaton grid is assigned a state based on these probabilities. For each cell, we generate a random number $\eta$ uniformly from $[0,1]$. We then determine the state $k$ of the cell by finding the smallest $k$ such that:

\[
\eta < \sum_{i=0}^k \Delta p_i.
\]

\subsubsection{Game-State-2 (GS2) Generation}\label{sec:gs2g}

GS2s were generated by applying a deterministic rules matrix (RM) to each cell in an initial configuration (IC). The total number of \emph{metastates} was denoted by \(c\). For every cell \(S_{(i,j)}\), the following procedure was carried out:

\begin{enumerate}
    \item \textbf{Metastate Identification.}  
    The metastate \(\mu_{(i,j)} \in \{1,\dots,c\}\) was determined by combining the current state of \(S_{(i,j)}\) with the number of its neighbors in state~1. This operation was count-based (isotropic), and \(\mu_{(i,j)}\) was defined as
    \[
    \mu_{(i,j)} \;=\; 1 \;+\; 
    \underbrace{\bigl(\text{current state of }S_{(i,j)}\bigr)\cdot 9}_{\substack{
    \text{shifts index by }9 \\
    \text{if cell state}=1}}
    \;+\;
    \underbrace{\bigl(\text{\# of neighbors in state }1\bigr)}_{\in\{0,\dots,8\}}.
    \]
    Thus, \(\mu_{(i,j)}\) took values in \(\{1,\dots,9\}\) if \(S_{(i,j)}\) was in state~0, or in \(\{10,\dots,18\}\) if \(S_{(i,j)}\) was in state~1.

    \item \textbf{Rule-Matrix Application.}  
    Each column of the RM, \(\text{RM}[\cdot,\mu_{(i,j)}]\), encoded a deterministic mapping from \(\mu_{(i,j)}\) to a single next state in \(\{0,1\}\). Because the RM was binary and deterministic, exactly one of its two rows was set to 1 for each column, and the other was 0. Concretely,
    \[
    \text{next state of }S_{(i,j)} 
    \;=\;
    \begin{cases}
    0 & \text{if }\text{RM}[0,\mu_{(i,j)}] = 1,\\
    1 & \text{if }\text{RM}[1,\mu_{(i,j)}] = 1.
    \end{cases}
    \]

    \item \textbf{Formation of the GS2 Configuration.}  
    Once the next state of every \(S_{(i,j)}\) was computed, these new states were assembled into a two-dimensional array of the same dimensions as the IC. This array was referred to as \(\text{GS2}\).
\end{enumerate}

Hence, \(\text{GS2}\) was generated by mapping each cell’s current state and neighborhood (its metastate) to a single next state through the RM. This process was repeated for all cells, yielding the complete evolved configuration after one timestep.

\subsubsection{Instruction Tuning}
\label{sec:instruction_tuning}

To incorporate the time-progression of each CA system into our training set, we represent the \emph{RM}, the \emph{IC}, and the resulting \emph{GS2} as a \emph{single string} with special tokens marking their boundaries. Concretely, after simulating the CA for one time-step (or more, depending on the experiment), we flatten each 2D matrix into a one-dimensional list and convert it to a space-delimited string. We then concatenate the three core elements (RM, IC, GS2) using reserved tokens as delimiters. An example format for the forward problem appears below:

\begin{center}
\verb+[BOS] [R] RM [BIC] IC [EIC] [BGS2] GS2 [EGS2] [EOS]+,
\end{center}

where:
\begin{itemize}
    \item \verb+[BOS]+ and \verb+[EOS]+ mark the start and end of the entire sequence.
    \item \verb+[R]+ introduces the flattened \emph{RM}.
    \item \verb+[BIC]+ (begin IC) and \verb+[EIC]+ (end IC) delimit the flattened \emph{IC}.
    \item \verb+[BGS2]+ (begin GS2) and \verb+[EGS2]+ (end GS2) delimit the \emph{GS2}.
\end{itemize}

 This preprocessed string is then tokenized (see section~\ref{sec:tok}) to form the input vector used for training.

\subsubsection{Data Splitting}
After a dataset of samples formatted for the relevant inference tasks of the corresponding model was stochastically generated, this dataset was shuffled using the \verb+.shuffle()+ function in the \verb+random+ python module. After shuffling, the initial 90\% of samples were assigned to the training dataset, the next 9\% to the validation dataset, and the final 1\% to the testing set.

\subsubsection{Tokenization}\label{sec:tok}
We employed a custom tokenizer based on the \verb+tokenizers+ python module~\cite{moi_huggingfaces_2023} that operates on strings of specific characters and groups of characters, separated by spaces. By predefining the vocabulary to only include necessary symbols for our specific type of data, we could drastically reduce the vocabulary size to only 22 unique tokens. We suspect that this small vocabulary size improved training efficiency compared to our previous LifeGPT model~\cite{berkovich_lifegpt_2024_updated} which used UTF-8-based tokenization. The tokenization process produced token indices that were subsequently transformed into vector representations via an embedding operation in our models.

\subsection{Inference}\label{sec:inference}
Inferences were performed in both forward and inverse modes to evaluate and validate model predictions under minimal prior assumptions. The procedure consisted of generating outputs from a pretrained model and comparing these outputs to ground truth data, either in the form of binary strings or rule matrices. The core methods are outlined below.

\paragraph{Forward Problem}
Forward inference was performed by iterating over a dataset of samples and extracting relevant segments from each sample. The model autoregressively generated an output string representing GS2 for every input. The following steps were used to compute a single accuracy metric per sample:
\begin{enumerate}
    \item \emph{Prompting and Inference.} Tokens corresponding to the ICs and RMs of testing set samples were extracted, tokenized, and provided to AutomataGPT. AutomataGPT then generated (with temperature set to 0) an inferred GS2, which was subsequently decoded into a human-readable string.
    \item \emph{Binary Extraction.} The decoded output was filtered to isolate only the binary data (\texttt{0} and \texttt{1}), and the ground truth underwent the same filtering process.
    \item \emph{Accuracy Computation.} A custom script was employed to compare predicted and ground-truth binary data, yielding an accuracy value for each sample.
    \item \emph{Accuracy Logging.} The per-sample accuracies were appended to a log file, and a running average was maintained throughout the process. Finally, after the last sample was evaluated, an average forward accuracy was computed across all samples.
\end{enumerate}

\subsubsection{Inverse Problem}
Inverse inference was the process of inferring an RM from an IC and GS2 pair, and validating the RM's accuracy by both comparing it with the ground truth RM and applying it to the sample IC to assess the accuracy of the resulting computed GS2. Each sample yielded two distinct accuracy metrics: one for the RM itself (\(A_{\text{RMI}}\)) and another for the application of the inferred RM to compute GS2 (\(A_{\text{IRMA}}\)). The procedure was as follows:
\begin{enumerate}
    \item \emph{Prompting and Inference.} Tokens corresponding to the ICs and GS2s of testing set samples were extracted, tokenized, and provided to AutomataGPT. AutomataGPT (with temperature set to 0) generated an inferred RM, which was decoded and reshaped into a \(\text{(2 \(\times\) 18)}\) array.
    \item \emph{RM Validation.} The predicted RM was checked to ensure that each column’s entries summed to~1, thereby confirming that valid transition probabilities were encoded. If the inferred RM was found to be illogical in this regard, both accuracy metrics (\(A_{\text{RMI}}\) and \(A_{\text{IRMA}}\)) were set to 0 for the corresponding sample.
    \item \emph{Evolution to GS2.} If validation succeeded, the IC was updated for one timestep using the inferred RM. The resulting configuration (computed GS2) was compared against the ground truth GS2 to compute the inferred RM application accuracy (\(A_{\text{IRMA}}\)). \(A_{\text{RMI}}\) was set to 1 if \(A_{\text{IRMA}}\) was also found to be 1 (indicating a degenerate solution), and otherwise it was set to the ratio of correctly inferred RM tokens to total RM length (see equation~\ref{eq:RMI}).
    \item \emph{Accuracy Logging.} Both \(A_{\text{RMI}}\) and \(A_{\text{IRMA}}\) were recorded for each sample in a log file. Average inverse problem accuracies were then computed (see equations~\ref{eq:AvRMI} and~\ref{eq:AvIRMA}).
\end{enumerate}

\subsection{Use of Generative AI}
Some Python scripts used for data generation, model training, data processing, and figure generation were written with the assistance of GPT-3.5, GPT-4, and GPT-4o from OpenAI. All scripts generated/edited in this manner were carefully reviewed, validated, and manually corrected, in the event of errors, by an author prior to implementation in our work.

\section*{Codes and Data}
All codes and data are available for non-commercial use at \url{https://github.com/lamm-mit/AutomataGPT}. 

\section*{Author Contributions}
JAB, NSD, and MJB conceived the concept, plan of study, developed the model and research, and wrote the paper. JAB developed the majority of algorithms, codes, and GitHub repository. JAB and NSD developed code for training data generation. JAB conducted the scientific investigations and data analysis. JAB and MJB revised and finalized the paper. 

\section*{Conflict of Interest}
JAB, NSD, and MJB declare inventorship on a U.S. provisional patent application filed by MIT (M.J. Buehler, J. Berkovich, N. David, Systems and Methods for Forecasting Using Cellular Automata, U.S. Provisional Application No. 63/825,225).

\section*{Acknowledgments}
JAB is supported by the Department of Defense (DoD) through the National Defense Science and Engineering Graduate (NDSEG) Fellowship Program.  


\bibliographystyle{unsrt}  
\bibliography{references,updatedrefs}

\end{document}